\documentclass[11pt]{article}

\usepackage[preprint]{acl}

\usepackage{times}
\usepackage{latexsym}

\usepackage[T1]{fontenc}

\usepackage[utf8]{inputenc}

\usepackage{microtype}

\usepackage{inconsolata}

\usepackage{graphicx}

\usepackage{longtable}
\usepackage{booktabs}
\usepackage{array}
\usepackage{pifont}
\usepackage{tcolorbox}
\usepackage{amsmath}

\title{ESF-Bench: Benchmarking Challenging Slot-Filling Scenarios\\for Real-World Enterprise Applications}

\author{Toby Liang \\
  ServiceNow \\
  \small \texttt{toby.liang@servicenow.com} \\\And
  Gopal Sarda \\
  ServiceNow \\
  \small \texttt{gopal.sarda@servicenow.com} \\\AND
  Sagar Davasam \\
  ServiceNow \\
  \small \texttt{sagar.davasam@servicenow.com} \\\And
  Vikas Yadav \\
  ServiceNow \\
  \small \texttt{vikas.yadav@servicenow.com} \\}

\begin{document}
\maketitle
\begin{abstract}
The rapid rise of large language models (LLMs) has driven transformative adoption across enterprises. However, deploying these models in real-world settings presents unique challenges due to complex system constraints and unexpected user behaviors. Among these applications, slot filling is essential for converting unstructured input into structured, actionable data. In this work, we introduce ESF-Bench, a challenging Enterprise Slot Filling benchmark consisting of 810 multi-turn samples and 6530 slots over 8 unique domains.  Curated using a taxonomy of the 57 most challenging slot-filling scenarios observed during real-world enterprise deployments, ESF-Bench exposes notable limitations in current state-of-the-art LLMs, with GPT-OSS-120b low successfully extracting slots for only 20.7\% of benchmark samples. To support continued research in this area, we publicly release the benchmark dataset, taxonomy, and accompanying evaluation code on GitHub\footnote{\url{https://github.com/ServiceNow/ESF-Bench}}.
\end{abstract}

\section{Introduction}
As large language models (LLMs) become increasingly integral to enterprise AI assistants, slot filling has emerged as a key capability for transforming unstructured user input into actionable data. Modern enterprise assistants often rely on slot filling to extract relevant information, such as account numbers, product categories, or scheduling details, from complex user-assistant conversations based on a task-oriented schema \cite{feng2023towards, xu2024large} . This structured data allows AI systems to automate business processes, enhance decision-making, and deliver personalized user experiences across multiple business domains such as IT service management, customer service management, and beyond \cite{bhargav2024approach}.

\begin{figure}[h]
    \centering
    \includegraphics[height=7cm, keepaspectratio]{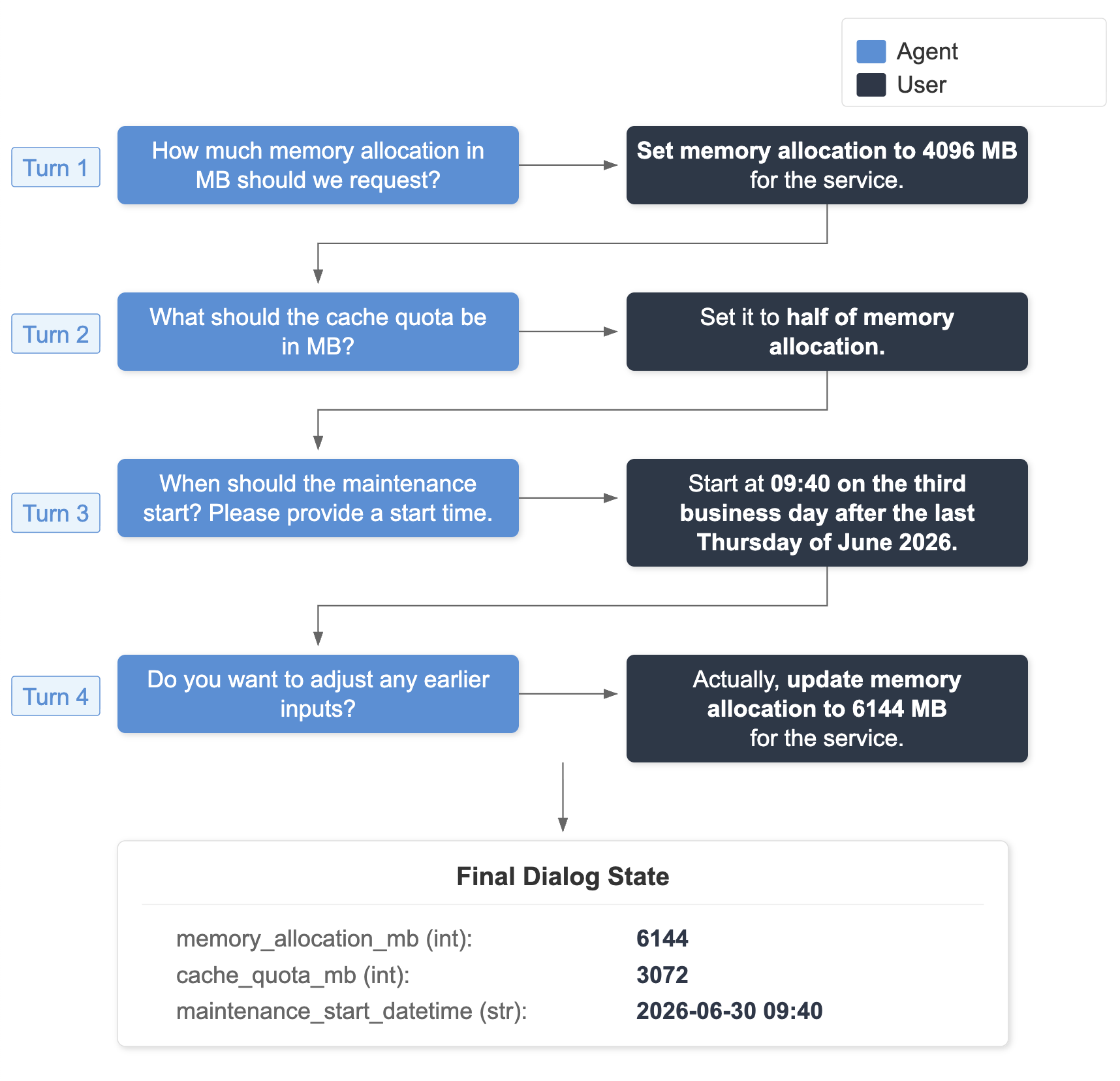}
    \caption{An example conversation and final dialog state from ESF-Bench. The user specifies slot values relative to other slots, later revises the original slot in a subsequent turn, and provides values that require complex temporal reasoning for accurate slot filling.}
    \label{fig:conversation}
\end{figure}

In practice, real-world applications of slot filling in enterprise contexts are often noisy and nuanced, requiring LLMs to handle a wide variety of scenarios where slot values exhibit intricate dependencies, constraints, and normalization requirements \cite{liu2021robustness, rana2025zero}.  Some applications require slots to be populated with multiple distinct values while others require slots to be extracted from various conflicting sources based on specialized priority and tiebreaker rules.  Meanwhile, users may exhibit unexpected behavior, such as providing information that is vague or contradictory, or by providing values that require mathematical, temporal reasoning, or counting abilities to resolve, leading to significant challenges in accurately extracting and populating the slots.

To systematically study these challenges, we introduce ESF-Bench, a comprehensive slot-filling benchmark comprising 6530 slots across 810 samples and spanning 8 enterprise domains including IT Service Management (ITSM), Customer Service (CSM), IT Operations Management (ITOM), Human Resources (HR), Finance, Medical, Law, and Education. Each sample is specifically designed to evaluate enterprise slot filling within a multi-turn user-assistant conversation and incorporates multi-source contextual metadata often found in real-world production environments such as user persona details, relevant knowledge-base (KB) articles, session information, and user-submitted forms, which must be leveraged to complete the slot-filling task effectively.

To curate the evaluation dataset, we first constructed a human-annotated taxonomy capturing the most common and complex features, constraints, and user behaviors.  Through careful observation of LLM performance in real-world slot filling systems, we identified 57 distinct scenarios to guide our data generation, ensuring both diversity and complexity in the resulting test set.  Tagging each slot with its corresponding scenario from the taxonomy enables fine-grained evaluation metrics that reveal model strengths and weaknesses at a much more granular level, allowing better diagnosis of performance gaps across various dimensions of the slot-filling task.

We summarize our contributions in the following manners:

1) First, we introduce ESF-Bench, a multi-domain slot-filling benchmark aimed at capturing the full complexity of real-world enterprise slot-filling systems, including multi-source context, intricate constraints, and unexpected user behaviors.

2) Additionally, we release a human-annotated taxonomy that formalizes the most difficult and commonly observed slot-filling scenarios in production, enabling fine-grained metrics that reveal specific strengths and weaknesses in model behavior.

3) We also provide an in-depth analysis of state-of-the-art LLMs on ESF-Bench, highlighting systematic capability gaps across taxonomy categories and offering insights that can guide the development of more robust and enterprise-ready slot-filling models.

\section{Related Work}
\label{sec:Related Work}
\subsection{Slot Filling Benchmarks for TOD Systems}
Research on \textbf{task-oriented dialog (TOD)} systems \cite{hosseini2020simple} has been largely focused on \textbf{dialog state tracking} \cite{jacqmin2022follow}, where models are evaluated on their ability to update dialog states through \textbf{natural language understanding (NLU)}, incrementally modifying intents and slots throughout the conversation. Prominent benchmarks such as \textbf{MultiWOZ} \cite{budzianowski2018multiwoz}, the \textbf{Schema-Guided Dialog (SGD)} dataset \cite{rastogi2020sgd}, and its extension \textbf{SGD-X} \cite{lee2021sgd-x} all follow this paradigm, providing annotations that support joint intent detection and slot filling at each dialog turn. These datasets typically rely on metrics such as \textbf{joint goal accuracy (JGA)} and \textbf{slot accuracy} \cite{jacqmin2022follow} to assess slot-filling performance.

More recently, ToD benchmarks have expanded their scope to multilingual settings. Datasets such as \textbf{BiToD} (English and Chinese) \cite{bitod2021} and \textbf{Multi3WOZ} \cite{Hu2023multi3woz} (4 languages), introduce multilingual slot-filling resources designed to support multilingual TOD applications.


\subsection{Single Utterance Slot Filling Benchmarks}
In parallel with advances in task-oriented dialog benchmarks, research on single-utterance slot filling has also been highly active, particularly within \textbf{spoken language understanding (SLU)}.  SLU focuses on mapping raw speech to intents and slots, and is commonly evaluated on benchmarks such as \textbf{ATIS} \cite{hemphill1990atis}, \textbf{SNIPS} \cite{coucke2018snips}, \textbf{Fluent Speech Commands} (FSC) \cite{lugosch2019speech}, and \textbf{SLURP} \cite{bastianelli2020slurp}, which provide audio-based annotations for assessing slot extraction directly from speech.  Another widely used benchmark is \textbf{MASSIVE} \cite{fitzgerald2023massive}, a text-based single-utterance slot-filling dataset derived from SLURP which covers 51 languages. Across these benchmarks, slot-filling accuracy and F1 scores are primarily used to evaluate single utterance slot-filling performance.

\subsection{Slot Filling Benchmarks for Enterprise}
Recently, efforts have increasingly focused on creating benchmarks that more accurately capture realistic enterprise domains and scenarios. \textbf{HR-MultiWOZ} \cite{xu2024hrmultiwoz}, for example, is an HR-focused TOD benchmark featuring purely extractive slots and long, domain-rich entities. It covers 10 distinct HR-related subdomains, including benefits enrollment, performance review, and relocation requests. Another benchmark, used as a held-out test set, encompasses over seven enterprise-oriented scenarios, introduces challenges such as long-form values, categorical slot decisions, and name-splitting tasks \cite{bhargav2024approach}. By partitioning test data across these scenarios, it enables fine-grained evaluation metrics tailored to each specific setting.

\section{ESF-Bench Overview}
Unlike the previous work discussed in Section \ref{sec:Related Work}, ESF-Bench is designed specifically to evaluate complex, unexpected, real-world slot-filling scenarios. While earlier benchmarks such as MultiWOZ \cite{budzianowski2018multiwoz} and SGD \cite{rastogi2020sgd} primarily emphasize extractive slots, ESF-Bench introduces a broader range of \textbf{abstractive cases} \cite{rana2025zero}, better reflecting production settings in which LLMs must normalize values, correct typos, or perform multi-step reasoning.

ESF-Bench also incorporates \textbf{multi-source} contextual metadata including user persona details, relevant knowledge-base articles, session information, and user-submitted forms. These additional sources are required for slots that depend on user personalization data or multi-hop reasoning to produce accurate predictions.

Because errors in intent detection can negatively affect slot-filling performance \cite{rana2025zero}, ESF-Bench \textbf{decouples slot filling from intent detection} where the input schema provides only the available slots. This separation enables a clearer evaluation of slot-extraction capabilities on their own, without relying on assumptions about upstream classification accuracy.


Finally, as ESF-Bench is based on an LLM’s ability to generate the complete and final dialog state, it \textbf{makes no assumptions on the downstream task}.  This perspective generalizes across evaluations of end-to-end TOD performance as well as offline tasks such as form population, case creation, and general conversational information extraction, independent of any particular dialog-management architectures or task formulations.

\subsection{Slot Filling Taxonomy}
\begin{figure*}[h]
    \centering
    \includegraphics[height=9cm, keepaspectratio]{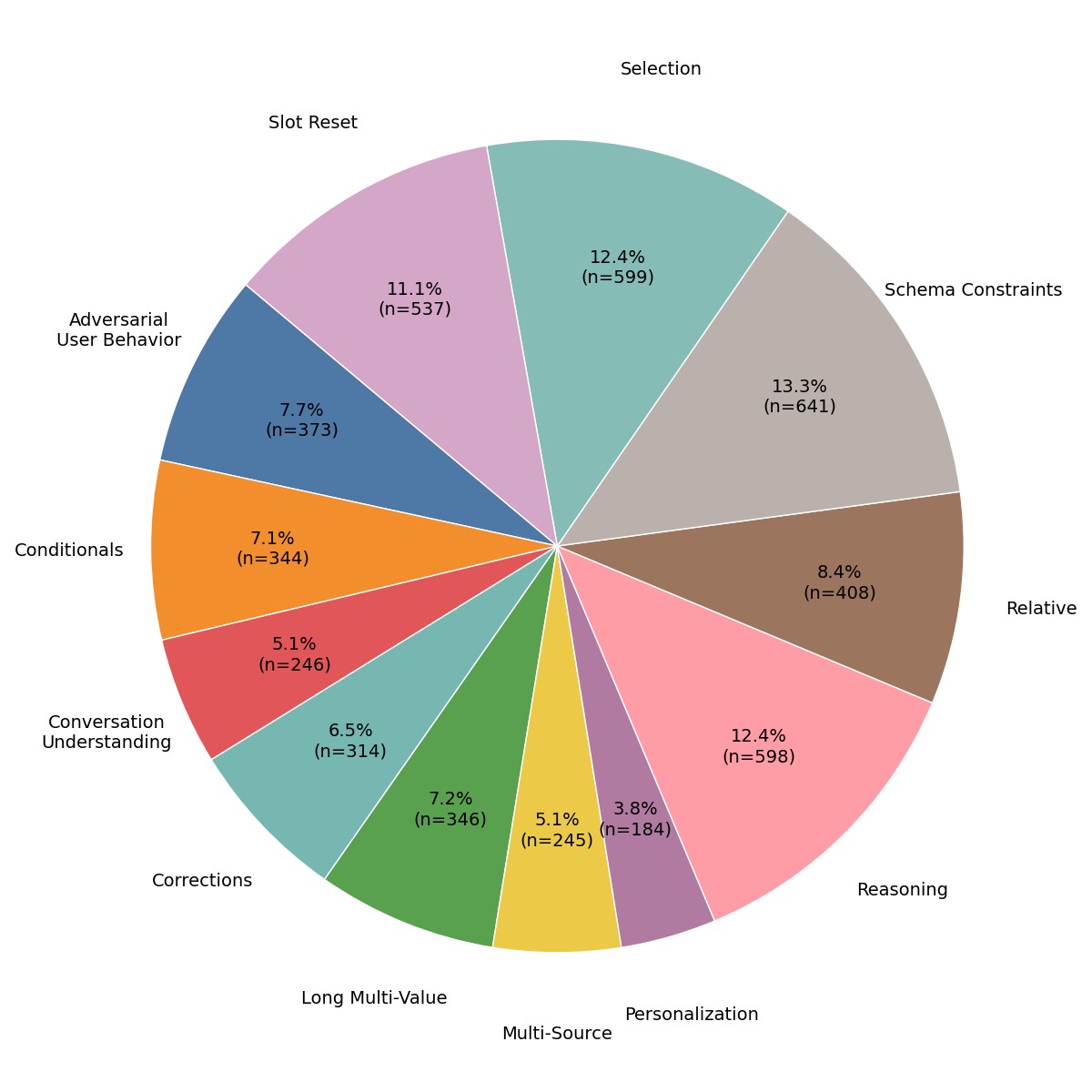}
    \caption{Distribution of ESF-Bench Taxonomy Categories}
    \label{fig:category_dist}
\end{figure*}


To generate ESF-Bench, we first constructed a taxonomy reflecting the most prevalent and challenging scenarios observed in production enterprise systems. We selected each scenario through detailed review of complex slot-filling prompts, defect reports, and LLM behavior in real-world workflows.

The resulting taxonomy comprises 57 scenarios, organized into 12 high-level categories: Selection (SEL), Conditionals (COND), Reasoning (REA), Multi-Source (MS), Long/Multi-Value (LMV), Conversation Understanding (CU), Personalization (PER), Schema Constraints (SC), Unexpected User Behavior (UUB), Relative (REL), Corrections (COR), and Slot Reset (SR). Figure \ref{fig:category_dist} shows the distribution of slots within each category and Table \ref{sec:taxonomy} in the appendix provides the full taxonomy.  The following sections examine the scenario types within each category.

\begin{table}[t]
\centering
\begin{tabular}{l c}
\toprule
\textbf{Statistic} & \textbf{Value} \\
\midrule
Total \# of Samples          & 810 \\
Total \# of Slots            & 6530 \\
Avg. \# Tokens per Prompt    & 4617.4 \\
Avg. \# Slots per Schema     & 8.06 \\
\# of Domains                & 8 \\
\# of Taxonomy Scenarios     & 57 \\
\bottomrule
\end{tabular}
\caption{Key Statistics of ESF-Bench}
\label{tab:esf_stats}
\end{table}

\subsubsection{Selection}
Often, multiple candidate values are available for a given slot. These candidates can originate from the conversation history or from external sources such as user persona data, knowledge base articles, session metadata, or user-submitted forms. Additionally, schemas may define a fixed set of valid slot values using an enum.

When multiple candidates exist, models must determine the most appropriate value based on the user's input. Users may provide information mapping directly to a defined enum value, or they may express preferences through comparisons (e.g., earlier than 9 am, cheaper than \$50) or superlatives (e.g., the fastest option, the second-cheapest plan). Values can also be inferred through negation (e.g., the no-meat option) or explicit specification (e.g., use my work email instead of my personal one). In certain cases, users may also select from a presented menu by referencing an option implicitly (e.g., option B) without stating the underlying value.

\subsubsection{Conditionals}
While users often specify slot values directly, they may also specify values through conditional statements. Specifically, users can supply an if-conditional (e.g., "If <condition> then...") or an if-else conditional (e.g., "If <condition> then ... else ...") to determine whether to fill a slot or which slot value to use. Conditionals may also govern whether to correct (Section \ref{sec:Corrections}) or reset (Section \ref{sec:Slot Reset}) a slot value. 
The schema itself may also include conditional statements that determine the correct default value or normalization rule to apply to a particular slot.

\subsubsection{Reasoning}
In the most challenging cases, slot filling requires advanced reasoning capabilities. This occurs when the correct slot value can only be derived through multi-step mathematical \cite{ahn2024large} or temporal reasoning \cite{chu2024timebench}, counting operations (e.g., counting the number of documents mentioned by the user), or general knowledge inference (e.g., inferring that the user is located in France when they mention Paris). In such situations, the necessary information may be distributed across the conversation history or drawn from multiple external sources. LLMs must therefore first retrieve all relevant context and then apply appropriate reasoning processes to infer the final slot value.

\subsubsection{Multi-Source}
Given multiple sources, such as user persona information, knowledge base articles, session metadata, and user-submitted forms, accurate slot filling often requires synthesizing information through multi-hop inference across these inputs. Furthermore, conflicting information may arise across sources due to outdated data, contextual differences, or retrieval errors. To address such conflicts, prompts may specify a source priority hierarchy, requiring LLMs to fill slot values based on the highest-priority source to determine the correct slot assignment.

\subsubsection{Long/Multi-Value}
Slot filling often involves extracting long or multi-value entries, particularly within the enterprise domain. For example, long identifiers, URLs, or serial numbers exceeding 512 characters are commonly found in enterprise data. Certain slots may also require multiple values to be filled, such as when a user specifies a list of permissions to enable. When values are excessively lengthy or numerous, LLMs must handle them with particular care and precision to ensure accurate extraction.

\subsubsection{Conversation Understanding}
In many scenarios, conversational understanding is critical for accurate slot assignment. Users may reference slot values discussed in previous turns without explicit repetition (e.g., let's go with the second option), requiring the model to maintain memory of prior alternatives. They may also propose slot values as rhetorical questions (e.g., how about 5pm?), and assistants may suggest values that users subsequently confirm. Finally, users may clarify or discuss the meaning of a potential slot value without actually committing to it. In all of these cases, LLMs must possess deep conversational understanding abilities to accurately assign the correct slot value.

\subsubsection{Personalization}
Personalization is an essential feature of enterprise solutions in which accurate slot value prediction often requires consideration of user persona information to deliver a tailored experience. Specifically, users may request slot values relative to their persona metadata (e.g., find me the option closest to my office). In adversarial scenarios where the retrieved user persona does not match the actual user (e.g., due to retrieval errors), LLMs must identify the persona mismatch and refrain from incorporating erroneous persona data into the slot filling task.

\subsubsection{Schema Constraints}
One of the most common cases involves constraints specified within the schema to determine whether or not to fill a slot. In particular, slots may require regex matching before extraction or impose semantic constraints based on general knowledge (e.g., the value must be a national capital city). Further, some slots may require units to be specified, while in other cases, pairs of slots may have cross-slot constraints (e.g., a check-out date must be after a check-in date). The schema may also specify particular methods for filling slots, such as providing default values when a slot is not explicitly specified or defining custom normalization rules to apply after extracting a slot value.

\subsubsection{Unexpected User Behavior}
In production scenarios, LLMs must be able to handle a variety of unexpected user behaviors \cite{liu2021robustness}. For example, users may specify slot values in ways that are uncertain, generic, or ambiguous, making it impossible to extract a single definitive slot value. Users may also specify values within hypothetical statements (e.g., If I were traveling for business, I'd book the Marriott, but this is personal) or embed them in sarcastic or non-serious comments. Additionally, users may make typo errors, produce confusing double-negation statements (e.g., It's not that I'm unopposed to using the backup server), or mention irrelevant third-party entities that should be ignored (e.g., Most people in sales use the CRM dashboard, but I need access to the analytics portal).

\subsubsection{Relative}
Rather than directly specifying a slot value, users may define values relative to other slots (e.g., Set the CPU limit to double the requested CPU). This relative specification creates dependencies between slots that must be maintained throughout the interaction. When a slot is defined relative to a previously filled slot, subsequent updates (see \ref{sec:Corrections}) to that base slot require implicit propagation of changes to all dependent slots. For example, if a user initially sets the CPU request to 2 cores and defines the CPU limit as double that value (4 cores), later updating the CPU request to 3 cores should automatically recalculate the CPU limit to 6 cores to preserve the established relationship.

\subsubsection{Corrections}
\label{sec:Corrections}
Often, users may revise slot values through corrections or updates. These corrections may specify explicit values or relative updates based on the original value (e.g., Actually, increase the headcount by 15\% and then add 5 more).  Users may also introduce conditional updates (e.g., If <condition>, then update <slot\_name> to <new\_value>) or reverse previous changes, restoring a slot to its original value. In adversarial scenarios, updated values may violate schema constraints, requiring the slot to be left unfilled.

\subsubsection{Slot Reset}
\label{sec:Slot Reset}
In addition to slot corrections, the user may opt to completely reset the slot value back to an unfilled state.  Similar to corrections, the user may specify a slot reset based on a condition or decide at a later time to undo the slot reset and keep the original slot value.  In special cases, users may request a full conversation reset, clearing all slot values and effectively restarting the conversation from its initial state.

\subsection{Benchmark Dataset Generation}
The ESF-Bench dataset comprises of 6530 slots over 810 samples. Each sample consists of a detailed prompt, multiple sources, and is selected from 1 of 8 domains. Comprehensive statistics for ESF-Bench are presented in Table~\ref{tab:esf_stats}.

\begin{table*}[t]
\centering
\resizebox{\textwidth}{!}{
\begin{tabular}{l c c c c c c c}
\toprule
Model & F-JGA & Tagged Slot Acc. & Median \# Total Tokens & Open Source & Reasoner \\
\midrule
Gemini 2.5 Flash Dynamic-Thinking & \textbf{33.3} & \textbf{80.9} & 3536.5 & \ding{55} & \ding{51} \\
GPT-5.1 High & \underline{30.7} & \underline{78.2} & 7611.5 & \ding{55} & \ding{51} \\
GPT-OSS-120b High & 23.8 & 76.5 & 5796.5 & \ding{51} & \ding{51} \\
Qwen3-32b-Think & 22.0 & 75.3 & 1656.5 & \ding{51} & \ding{51} \\
Ministral3 14b Reasoning & 20.9 & 74.1 & 10271.0 & \ding{51} & \ding{51} \\
\midrule
Gemini 2.5 Flash No-Thinking & 9.2 & 64.5 & 209.0 & \ding{55} & \ding{55} \\
GPT-5.1 Minimal & \underline{10.9} & \underline{67.3} & 179.0 & \ding{55} & \ding{55} \\
GPT-OSS-120b Low & \textbf{20.7} & \textbf{73.0} & 844.5 & \ding{51} & \ding{51} \\
Qwen3-32b-No-Think & 4.3 & 56.3 & 176.5 & \ding{51} & \ding{55} \\
Ministral3 14b Inst & 3.7 & 56.7 & 194.5 & \ding{51} & \ding{55} \\
\midrule
High-Capability Avg. & 26.1 & 77.0 & 5774.4 & - & - \\
Low-Latency Avg. & 9.8 & 63.6 & 320.7 & - & - \\
\bottomrule
\end{tabular}
}
\caption{Evaluation of high-capability and low-latency large language models on ESF-Bench. Models with extended reasoning (top group) significantly outperform non/low-reasoning models (bottom group) on F-JGA and tagged slot accuracy, though at higher token costs.}
\label{tab:overall_results_table}
\end{table*}

\begin{figure}[h]
    \centering
    \includegraphics[width=\columnwidth]{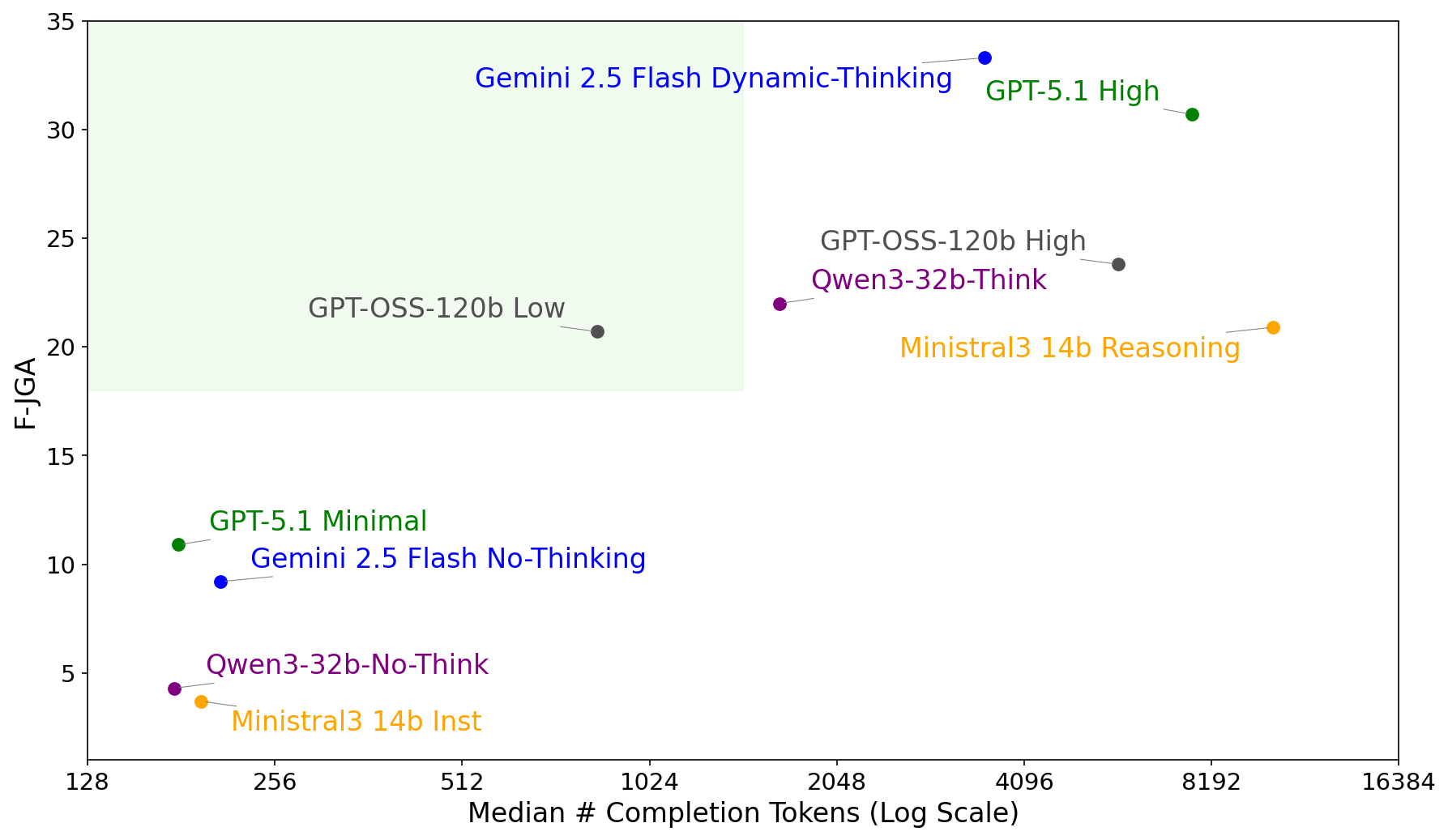}
    \caption{Performance–efficiency trade-off across SOTA models. F-JGA is plotted against median completion length on a log scale.}
    \label{fig:quadrant_plot}
\end{figure}

\subsubsection{Data Synthesis Methodology}
Building on LLM-driven synthetic data generation \cite{long2024llms} and LLM-based evaluation \cite{zheng2023judging}, we developed an in-house data synthesis engine that converts our taxonomy of enterprise slot-filling scenarios into benchmarking samples. The pipeline begins by stochastically sampling slot-filling scenarios from the taxonomy, which are then used to systematically generate prompt components and their corresponding metadata through a series of LLM calls. Ground truths are then generated using a separate LLM call conditioned on the same metadata used for prompt generation, with a judge model verifying label correctness. Each slot in the dataset is tagged with its source scenario from the taxonomy, enabling fine-grained performance analysis across different enterprise use cases.  Within our pipeline, we use GPT-5 high both for data generation and judging and GPT-4.1 for scenario tagging.  To validate the quality of the synthetically generated data, we conducted a human annotation exercise on a representative sample of 122 tagged slots where human annotators achieved a Tagged Slot Accuracy of 93.83\%. We report additional human performance results in section \ref{sec:human_perf_baseline}. An example prompt can be found in section \ref{sec:prompt_example} of the appendix.

\subsubsection{Multiple Sources}
Each prompt within our dataset includes a synthetically generated task-oriented schema, user-assistant conversation, and auxiliary sources such as knowledge base articles, user persona details, session information, and previous user form submissions. These sources model the metadata typically found in real enterprise applications. Samples in ESF-bench are specifically generated to require cross-referencing information from these sources to the conversation history to accurately complete the slot filling task.

\subsubsection{Output Format Conversion}
As there is no standard output format for slot filling, we evaluate model robustness by varying the output format specifications across prompts.  While this approach enhances prompt diversity, it complicates the evaluation process when comparing model predictions against ground truth, as each output adheres to a different structure.

To address this challenge, we employ an LLM to automatically generate a Python function that maps each prompt's output to a standardized format suitable for evaluation. This strategy enables us to preserve the diversity of our prompt set while standardizing comparisons across all model outputs.

\begin{table*}[t]
\centering
\resizebox{\textwidth}{!}{
\begin{tabular}{l l l l l l l l l l l l l}
\midrule
Model & UUB & COND & CU & COR & LMV & MS & PER & REA & REL & SC & SEL & SR \\
\midrule
Gemini 2.5 Flash Dynamic-Thinking & \textbf{74.5} & \underline{88.1} & 89.0 & \textbf{84.1} & 77.7 & \underline{86.9} & 58.7 & \underline{64.0} & \underline{85.0} & \underline{82.5} & \textbf{89.5} & \textbf{90.1} \\
GPT-5.1 High & \underline{64.9} & \textbf{88.4} & \textbf{93.1} & 78.3 & \underline{78.6} & 84.1 & 47.3 & \textbf{64.2} & 79.4 & 80.7 & 82.6 & \underline{88.6} \\
GPT-OSS-120b High & 59.5 & 87.2 & \underline{89.8} & 72.6 & \textbf{81.5} & 80.8 & 45.7 & 60.4 & 76.0 & \textbf{83.8 }& 80.3 & 87.0 \\
Qwen3-32b-Think & 57.9 & 75.0 & 86.6 & \underline{83.4} & 74.3 & \textbf{87.3} & \textbf{66.8} & 53.8 & \textbf{91.2} & 74.6 & \underline{83.1} & 82.5 \\
Ministral3 14b Reasoning & 60.1 & 74.4 & 88.6 & 76.8 & 75.7 & 78.4 & \underline{60.9} & 54.5 & 78.7 & 77.1 & 80.8 & 84.4 \\
\midrule
Gemini 2.5 Flash No-Thinking & \textbf{66.8} & 60.7 & 76.5 & \underline{68.1} & 63.5 & 71.9 & \underline{62.3} & 22.1 & \underline{66.2} & 70.5 & 72.3 & 79.4 \\
GPT-5.1 Minimal & \underline{62.5} & \underline{64.2} & \underline{82.5} & 67.8 & \underline{68.5} & \underline{73.1} & \textbf{64.7} & \underline{23.4} & 65.4 & \underline{77.1} & \underline{75.6} & \textbf{86.0} \\
GPT-OSS-120b Low & 57.6 & \textbf{85.5} & \textbf{86.6} & \textbf{70.7} & \textbf{76.3} & \textbf{75.9} & 42.4 & \textbf{52.7} & \textbf{72.1} & \textbf{82.4} & \textbf{81.5} & \underline{80.4} \\
Qwen3-32b-No-Think & 50.9 & 59.0 & 61.8 & 64.6 & 59.2 & 66.1 & 58.2 & 20.6 & 56.9 & 62.1 & 64.8 & 64.4 \\
Ministral3 14b Inst & 58.2 & 54.7 & 69.9 & 59.6 & 52.6 & 66.9 & 55.4 & 17.6 & 49.8 & 64.9 & 68.6 & 64.8 \\
\midrule
High-Capability Avg. & 63.4 & 82.6 & 89.4 & 79.0 & 77.6 & 83.5 & 55.9 & 59.4 & 82.1 & 79.7 & 83.3 & 86.5 \\
Low-Latency Avg. & 59.2 & 64.8 & 75.5 & 66.2 & 64.0 & 70.8 & 56.6 & 27.3 & 62.1 & 71.4 & 72.6 & 75.0 \\
\bottomrule
\end{tabular}
}
\caption{Slot-level performance breakdown by taxonomy category. Models with extended reasoning (top group) demonstrate superior handling of complex slot filling scenarios including relative (REL), conditionals (COND), and reasoning (REA).}
\label{tab:fine_grained_results_table}
\end{table*}

\subsubsection{Dataset Domains}
To ensure domain diversity, our dataset spans over 8 enterprise domains: IT Service Management (ITSM), Customer Service Management (CSM), IT Operations Management (ITOM), Human Resources (HR), medical, legal, finance, and education.  For each domain, we compiled 25 specific instances (e.g., password reset for ITSM) that are randomly sampled during prompt generation. This approach yields 200 unique domain-instance pairs, enabling diverse scenario specification throughout the dataset.

\section{Experimental Setup}

\subsection{Model Selection}
We select five model families to conduct our experiments (GPT-5.1, Gemini 2.5 Flash \cite{comanici2025gemini}, GPT-OSS-120b \cite{agarwal2025gptoss}, Qwen3-32b \cite{qwen2025qwen3}, and Ministral3 14b \cite{mistralai2025ministral}), and evaluate each under two deployment configurations. The \textbf{High-Capability models} includes the reasoning-optimized variant, leveraging extended reasoning through increased test-time compute \cite{snell2025scaling,deepseek2025r1} to target scenarios where quality is prioritized over latency. The \textbf{Low-Latency models} includes the corresponding efficient variant of each family reflecting production deployments where token consumption and response latency are primary constraints. Notably, GPT-OSS-120b Low retains reasoning capabilities despite being in the low-latency tier, highlighting that reasoning ability and latency efficiency are not mutually exclusive. We include both proprietary and open-source models to represent realistic deployment scenarios where organizations commonly leverage a mix of commercial and open-source solutions.

\begin{figure*}[t]
  \centering
  \includegraphics[width=0.48\linewidth]{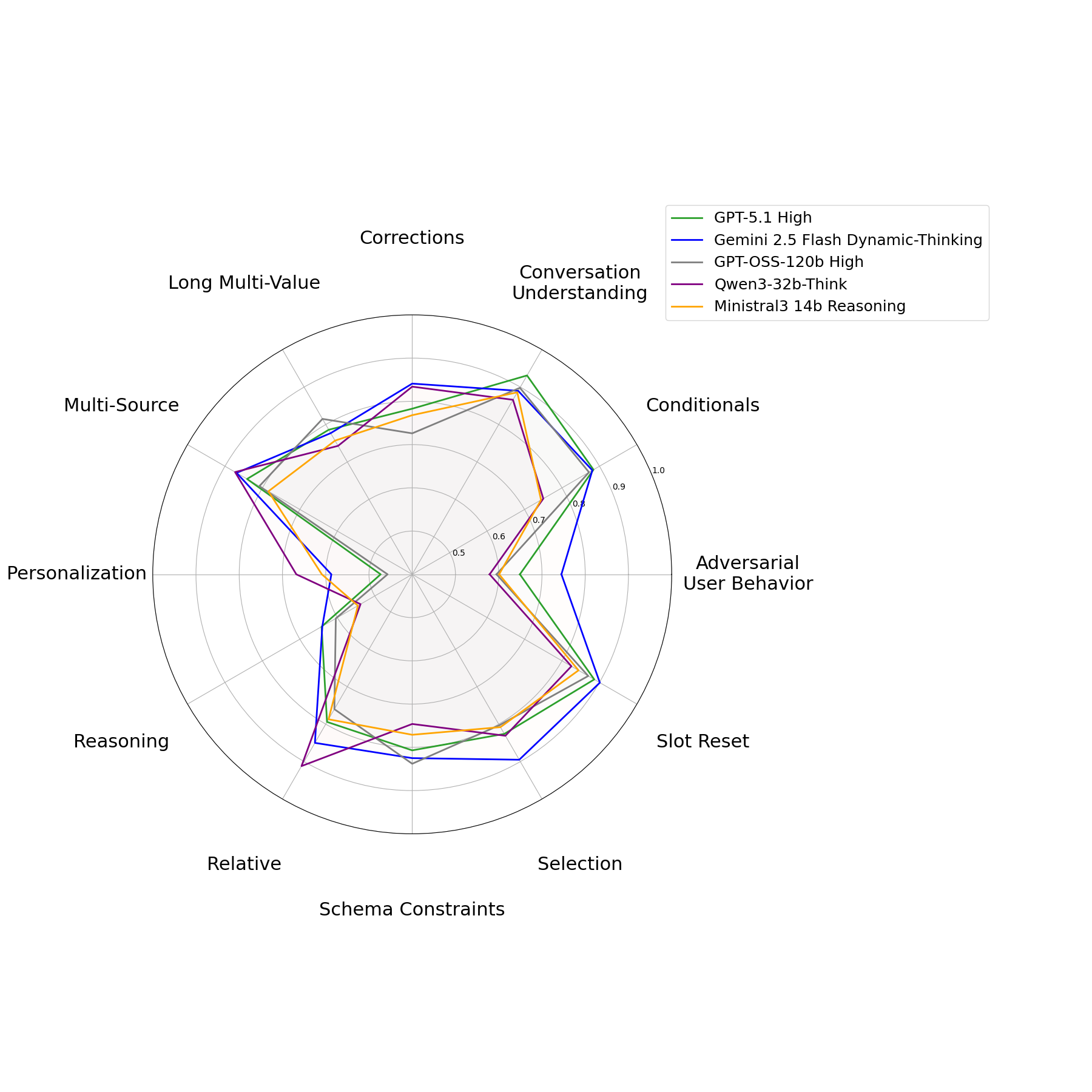}
  \includegraphics[width=0.48\linewidth]{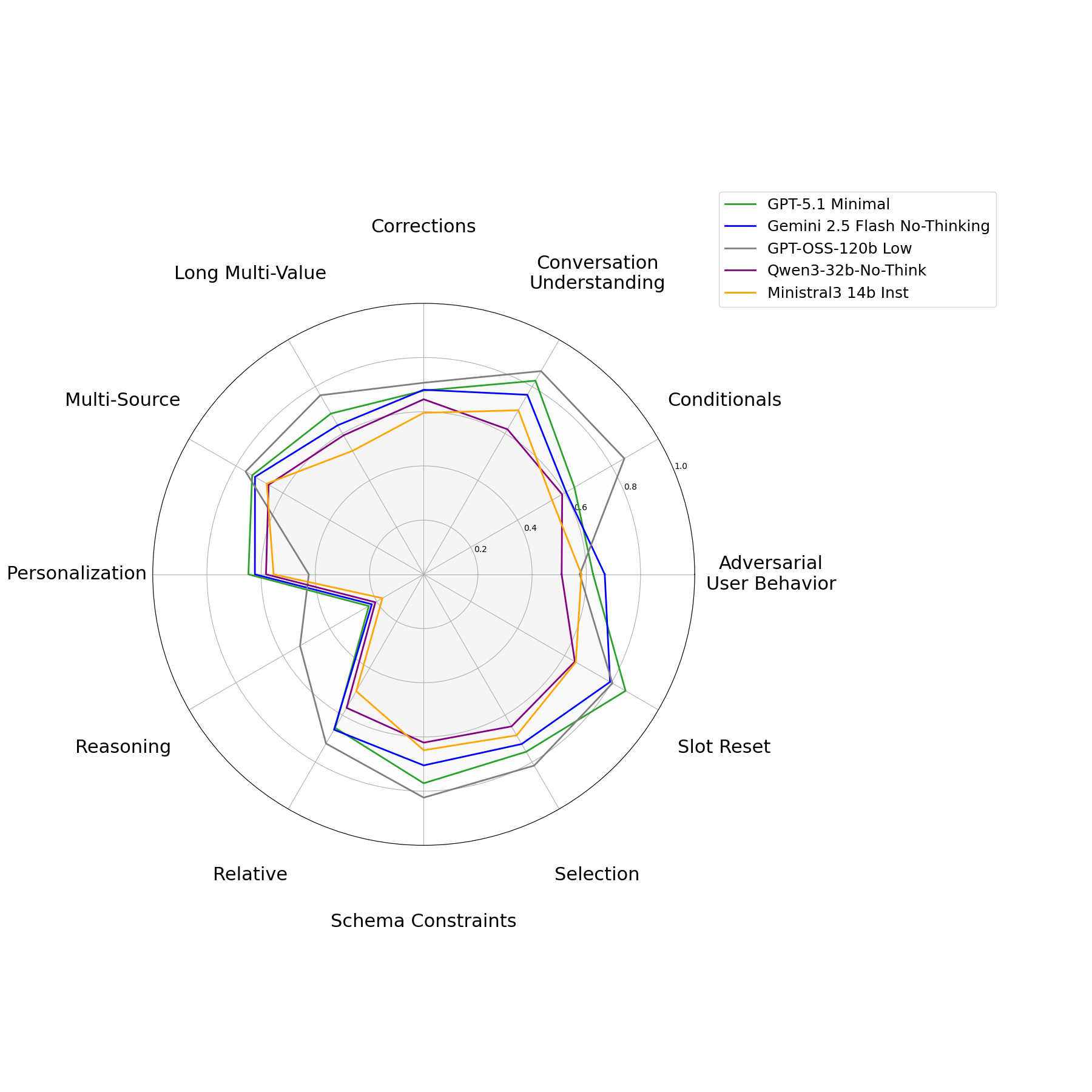}
  \caption{Slot-Level Category Scores For High-Capability Models (Left) and Low-Latency Models (Right)}
\end{figure*}

\subsubsection{High-Capability Models}
\label{sec:Reasoning Models}
To evaluate slot filling performance on ESF bench, we configure each reasoning model with its maximum reasoning capacity, setting GPT-5.1 and GPT-OSS to "high" reasoning effort, Gemini 2.5 Flash to "dynamic thinking" mode, Qwen3 32b to "think mode," and deploy the reasoning variant of Ministral3 14b. Across all models, we set max tokens to 32k and temperature to 0.6.

\subsubsection{Low-Latency Models}
For each model specified in Section~\ref{sec:Reasoning Models}, we evaluate a corresponding non-reasoning or low-reasoning variant, disabling thinking entirely for GPT-5.1, Gemini 2.5 Flash, and Qwen3 32b, while GPT-OSS-120B is evaluated with reasoning effort set to low. For Ministral3 14b, we use the instruct version as the non-reasoning baseline. We configure these experiments with a maximum token limit of 32k and set the temperature to 0.6 for GPT-OSS-120B (low reasoning) and 0.001 for all other models.

\subsection{Metrics Definitions}
Traditional dialogue evaluation metrics like Joint Goal Accuracy (JGA) \cite{jacqmin2022follow} measure state tracking at individual turns, limiting their applicability to task-agnostic settings where the complete dialogue outcome matters most. Since ESF-bench is designed to generalize across both online task-oriented dialogue (TOD) systems and offline evaluation tasks, we require a metric that evaluates performance based on the complete final dialog state rather than turn-by-turn accuracy.  We therefore introduce \textbf{Final Joint Goal Accuracy (F-JGA)}, which extends JGA by evaluating whether all slots in the final dialogue state are correct (see Appendix~\ref{app:fjga} for formal definition). This holistic metric directly assesses whether a model can comprehend the full conversational context and reliably extract all required slots simultaneously, both critical capabilities for real-world dialogue systems where partial solutions often provide little value.

In addition to F-JGA, we measure the accuracy of slots tagged with at least one scenario from our taxonomy, which represents 66.8\% of all slots. While untagged slots still contribute to the F-JGA score, they can often be filled through simple extraction and provide limited insight when evaluating slot-filling capabilities. Therefore, we compute \textbf{slot accuracy} exclusively for tagged slots to better assess the model's performance on these more challenging scenarios.

Finally, we report the \textbf{median number of tokens} per response, including reasoning traces. We use the median rather than the mean because reasoning trace lengths often contain outliers that would significantly skew the average, making the median a more robust measure of typical model behavior.

\section{Results}

We present our findings in two parts.  First we report overall F-JGA, tagged slot accuracy, and median number of total tokens under Table \ref{tab:overall_results_table}.  We also report fine-grained category-level slot accuracy metrics under Table \ref{tab:fine_grained_results_table}.

Overall, we find that while slot filling is a well-known task, certain aspects of slot filling still remain a challenge, with certain models performing better or worse based on specific scenarios or constraints.  Among the evaluated models, Gemini2.5 Flash Dynamic-Thinking achieves the strongest overall performance, achieving the best results when dealing with comparisons (98.0\%), superlatives (89.6\%), and negations (82.2\%).

Across all evaluated taxonomy categories, we find that reasoning and personalization presents the greatest challenge for the highest-performing LLMs, consistent with known difficulties in mathematical \cite{ahn2024large} and temporal reasoning \cite{chu2024timebench}, with high capability models scoring only 59.4\% on the reasoning category and just 55.9\% on personalization. In contrast, models generally demonstrate strong performance on conversation understanding, averaging 89.4\% across high-capability models and 75.5\% even among low-latency models.


\begin{table}[t]
\centering
\small
\begin{tabular}{@{}lcc@{}}
\toprule
\textbf{Model} & \textbf{F-JGA} & \textbf{Tagged Slot Acc.} \\
\midrule
Human & 73.33 & 93.83 \\
\midrule
Gemini 2.5 Flash DT & 40.0 & 82.7 \\
GPT-5.1 High & 33.3 & 76.5 \\
\bottomrule
\end{tabular}
\caption{Human performance baseline compared to top-performing models on a representative subset (122 tagged slots). DT = Dynamic-Thinking.}
\label{tab:human_baseline}
\end{table}

\subsection{High-Capability vs Low-Latency Models}
High-capability (reasoning-optimized) models demonstrate substantial advantages over low-latency (low/non-reasoning) models across both metrics, achieving average improvements of +16.3\% in F-JGA and +13.4\% in tagged slot accuracy. These gains extend across nearly all fine-grained taxonomy categories, with the most improvements observed in reasoning (+32.1\%), relative (+20\%), and conditionals (+17.8\%) categories.

We find that high-capability models perform better on tasks requiring multiple steps, particularly in scenarios where logical inference or calculations are necessary for accurate slot filling. This is most evident in math (+44.0\%), temporal (+33.2\%), and counting (+32.0\%) categories, where producing the correct value requires multi-step inference that low-latency models consistently fail to carry out.

Relative categories such as relative corrections (+54.7\%) and relative slot (+43.9\%) show similarly large disparities, as low-latency models often fail to remember or aggregate information across multiple conversation turns before deriving the final slot value. High-capability models also outperform low-latency models in conditional cases like conditional counting (+29.6\%), conditional corrections (+21.5\%), and conditional reset (+20.6\%), where low-latency models struggle to correctly evaluate the condition before determining the correct slot to populate.

On the other hand, our experiments also reveal specific failure cases for high-capability models, particularly when handling unexpected user behavior. For third-party entity references, high-capability LLMs fail to recognize that a mentioned entity belongs to a third party and incorrectly extract slot values as if they were the user's own. Similarly, when users express preferences sarcastically with exaggerated or implausible values, models often take these statements at face value and extract the slot accordingly. Double negations also pose a challenge, as models become confused by the layered negation structure and frequently misinterpret the user's actual intent.

In the personalization category, while high-capability models outperform low-latency models on standard cases where the user persona matches the user within conversation history, they frequently fail in adversarial scenarios featuring user persona mismatches, often resulting in invalid slot fills.  Further, we continue to notice some cases where models extract slots values that require units, even when units are not explicitly provided in the context.

\subsection{Open Source vs Closed Models}
We find that among the evaluated models, open source models still provide a viable alternative in certain areas.  GPT-OSS-120b High scores within -4.4\% tagged slot accuracy of the best proprietary model and leads on long/multi-value and schema constraints, while Qwen3-32b-Think tops all models on multi-source, personalization, and relative slots, though both trail proprietary models on areas like unexpected user behavior, reasoning, and overall F-JGA.

\subsection{Human Performance Baseline}
\label{sec:human_perf_baseline}
To contextualize model performance, we established a human baseline by having annotators complete the slot-filling task on a representative subset of 122 tagged slots. As shown in Table \ref{tab:human_baseline}, human annotators achieved an F-JGA of 73.33 and a Tagged Slot Accuracy of 93.83\%, substantially outperforming even the best model (Gemini 2.5 Flash Dynamic-Thinking at 40.0 F-JGA). The 33.33-point gap in F-JGA between human and model performance demonstrates that ESF-Bench remains a non-trivial and challenging benchmark for current state-of-the-art LLMs. Notably, the gap narrows at the slot level (93.83\% vs 82.7\%), suggesting that while models can handle individual slots reasonably well, they struggle to get \textit{all} slots correct simultaneously within a single sample, a requirement that is critical for real-world enterprise deployments where partial solutions often provide little value.

\section{Conclusion}
In this paper, we present ESF-Bench, a benchmark designed to reflect the noisy, constrained realities of enterprise slot-filling systems. Our empirical evaluations reveal scenarios where even state-of-the-art LLMs fall short, while highlighting the performance-latency tradeoffs between full reasoning models and their efficient variants. By releasing ESF-Bench with its associated data, code, and taxonomy, we aim to provide the community with a foundation for systematically advancing more reliable and enterprise-ready slot-filling systems.

\section{Limitations}


\subsection{Error Attribution in Slot Prediction}
While each slot is tagged with its associated taxonomy scenario, determining the cause of a mispredicted slot still requires manual review of the sample.  For example, a successful slot extraction might require both correct extraction of a slot as well as applying the proper normalization rule.  However, it is non-trivial to determine which step a particular model failed at.  Therefore, future work can be done to automate the process of determining the exact cause of mispredictions.

\subsection{Synthetic Data Generation Bias}
As ESF-Bench is generated using a fully synthetic data generation pipeline based on multiple LLM calls, there may exist the possiblity of evaluation biases.  These biases could be more pronounced for models from the same family or for models that share training data with those used to generate the dataset (GPT-5 and GPT-4.1). Given this, our results show that GPT-based models do not score significantly higher than other models. Nonetheless, further analysis can be done to better understand and quantify the potential impact of benchmark bias in synthetically generated data.

\subsection{Language and Modalities Support}
Though all slots and sources are primarily in English, support for multiple languages may be greatly beneficial to global enterprises spanning multiple countries and regions.  Similarly, slot-filling can be expanded to include multi-modal support where user and assistant conversations may be provided through audio recordings and auxilary sources may contain images of tables and figures where slots can be extracted from.  To this end, future work can be done to extend ESF-Bench support to multilingual and multimodal domains.

\bibliography{custom}

\clearpage
\onecolumn
\appendix
\section*{Appendix}
\section{ESF-Bench Taxonomy}
\label{sec:taxonomy}
\begin{center}
\renewcommand{\arraystretch}{1.35}

\begin{longtable}{p{0.3\textwidth} p{0.15\textwidth} p{0.5\textwidth}}

\hline
\textbf{Scenario} & \textbf{Categories} & \textbf{Description} \\
\hline
\endfirsthead

\hline
\textbf{Scenario} & \textbf{Categories} & \textbf{Description} \\
\hline
\endhead

\multicolumn{3}{r}{\textit{Continued on next page}} \\
\hline
\endfoot

\endlastfoot
Valid Enum & Selection & Schema provides a list of enums for a particular slot and the user specifies a slot values that maps to an enum. \\ \hline

Invalid Enum & Selection & Schema provides a list of enums for a particular slot and the user specifies a slot values that does not map to any of the enums. \\ \hline

Comparison & Selection & Multiple options for a slot exist throughout the conversation and sources.  The user specifies a slot using a comparison statement (e.g., Find me an option that arrives before 5pm.). \\ \hline

Adversarial Comparison & Selection & Multiple options for a slot exist throughout the conversation and sources.  The user specifies a slot using a comparison statement but none of the options satisfy the statement. \\ \hline

Superlative & Selection & Multiple options for a slot exist throughout the conversation and sources.  The user specifies a slot using a superlative statement (e.g., Find me the option that is least expensive after applying state tax.). \\ \hline

Negation & Selection & Multiple options for a slot exist throughout the conversation and sources.  The user specifies a slot using a negation statement (e.g., I don't want any options that contain meat). \\ \hline

Adversarial Negation & Selection & Multiple options for a slot exist throughout the conversation and sources.  The user specifies a slot using a negation statement but multiple options fit the criteria. \\ \hline

Ambiguous Slot Specification & Selection & Multiple conflicting slot values are present within various sources and the user specifies one of them (e.g., Use my work email instead of my personal one). \\ \hline

Conversation History Menus & Selection & The assistant provides a menu of options to choose from (a), (b), (c), or (d) and the user selects from on or more of the provided options. \\ \hline

If Conditionals & Conditionals & User specifies a slot value based on an if condition (e.g., If <condition>, then <slot> should be <value>). \\ \hline

If Else Conditionals & Conditionals & User specifies a slot value based on an if/else condition (e.g., If <condition>, then <slot> should be <value1>.  Otherwise, it should be <value2>). \\ \hline

Math & Reasoning & Requires determining relevant numerical information and multiple steps of arithmetic to fill the slot. \\ \hline

Temporal & Reasoning & Requires temporal reasoning based on dates and time to determine the slot. \\ \hline

Counting & Reasoning & Requires counting occurrences of relevant information to fill the slot.   \\ \hline

Conditional Counting & Reasoning, Conditionals & Requires counting occurrences of relevant information based on conditions (e.g., Only count if <condition>) to fill the slot. \\ \hline

General Knowledge Inference & Reasoning & Requires slots to be filled using general knowledge inference (e.g., If user mentions apt-get, primary\_os\_family should be set to Linux). \\ \hline

Multi-Hop & Multi-Source & Requires synthesis of information from multiple sources and multi-hop inference to fill the slot. \\ \hline

Source Priority & Multi-Source & Conflicting information is found in multiple sources and source priority and tiebreaker instructions are required to determine the correct slot value to fill. \\ \hline

Long Value & Long/Multi-Value & Slot values are provided with extremely long values (64+ characters) such as IDs, URLs, or tokens. \\ \hline

Multi Value Slots & Long/Multi-Value & Slots requiring multiple values within a single slot. \\ \hline

Multi Value Slots Constraints & Long/Multi-Value & Slots requiring multiple values within a single slot filtered by specific constraints defined within the schema. \\ \hline

Conversation Memory & Conversation Understanding & Assistant provides options in a previous turn and the user selects one of the options later in the conversation (e.g., 5 more than the first option we spoke about previously.) \\ \hline

Rhetorical Question & Conversation Understanding & User specifies a slot value using a rhetorical question (e.g., How about 5pm?). \\ \hline

Assistant Proposal & Conversation Understanding & The slot value is proposed by the assistant and confirmed by the user rather than the user specifying the slot value directly. \\ \hline

Adversarial Assistant Proposal & Conversation Understanding & The slot value is proposed by the assistant and confirmed by the user but the value proposed by the assistant is invalid based on schema constraints. \\ \hline

Meta Discussion & Conversation Understanding & User clarifies and discusses the meaning of a possible slot value with the assistant without actually confirming the slot value. \\ \hline

User Personalization & Personalization & Slot filling requires custom personalization based on user persona metadata (e.g., find me the option that is closest to my office). \\ \hline

Adversarial User Persona & Personalization & The user within the conversation does not match the user persona metadata.  Only use information provided directly from the user.\\ \hline

Adversarial User Persona No Value & Personalization & The user within the conversation does not match the user persona metadata.  The user does not provide any information and the user persona metadata should be ignored as well. \\ \hline

Default Value & Schema Constraints & The schema specifies a default value to use when no slot value can be extracted. \\ \hline

Conditional Default Value & Schema Constraints, Conditionals & The schema specifies a default value based on a condition (e.g., If <condition>, then use <default\_value$_1$> otherwise use <default\_value$_2$>.\\ \hline

Normalization & Schema Constraints & The schema contains custom normalization rules to use to extract a slot into a standardized format. \\ \hline

Conditional Normalization & Schema Constraints, Conditionals & The schema specifies a normalization rule based on a condition (e.g., If <condition>, then normalize to <normalization\_rule$_1$>, otherwise use <normalization\_rule$2$>). \\ \hline

Syntax Constraints & Schema Constraints & The schema contains syntax constraints that must be adhered to before a slot can be extracted (e.g., contains numeric digits, avoid specific keywords, is all upper-case, etc.) \\ \hline

General Knowledge Constraints & Schema Constraints & The schema mentions constraints that require general or common knowledge to resolve.  (e.g., must be a national captital city).\\ \hline

Units Required & Schema Constraints & The schema provides a slot that requires a specific unit (e.g., attendance\_duration\_minutes) but the user does not specify any units. \\ \hline

Cross Slot Constraints & Schema Constraints & The schema specifies constraints between a pair of slots (e.g., check\_out\_date > check\_in\_date). \\ \hline

Cross Slot Constraints Corrections & Schema Constraints & The schema specifies constraints between a pair of slots (e.g., check\_out\_date > check\_in\_date).  Later in the conversation, a correction is made to one of the slot values to validate or invalidate this constraint. \\ \hline

Third Party Entity No Value & Unexpected User Behavior & User mentions a slot value from a third party entity without ever providing their own value (e.g., My spouse uses Brookside Pharmacy on Pine). \\ \hline

Double Negation & Unexpected User Behavior & User specifies a slot value using double negation (e.g., I don't mean that I don't want to not have the account frozen). \\ \hline

Uncertainty & Unexpected User Behavior & User provides a slot value in an uncertain manner (e.g., Maybe around 3-ish but I'm not sure). \\ \hline

Generic & Unexpected User Behavior & User provides a slot value in a vague manner (e.g., Let's do it sometime after quarter end.). \\ \hline

Ambiguous & Unexpected User Behavior & The user specifies multiple values (e.g., primary vehicle, secondary vehicle, rental vehicle, etc.) for a single value slot. \\ \hline

Sarcasm & Unexpected User Behavior & User specifies a slot value in a sarcastic or non-serious tone and the slot should not be extracted. \\ \hline

Typo & Unexpected User Behavior & The user makes a typo mistake for common English words within the slot value. \\ \hline

Relative Slot & Relative & Slots are specified relative to one of the previously filled slots (e.g., Make the CPU limit equal to double the requested CPU). \\ \hline

Relative Slot Correction & Relative, Corrections & Slots are specified relative to one of the previously filled slots.  Later in the conversation, the previously filled slot is updated so that the relative slot must be implicitly updated too. \\ \hline

Relative Slot Reset & Slot Reset, Relative & Slots are specified relative to one of the previously filled slots.  Later in the conversation, the previously filled slot is reset so that the relative slot must be implicitly reset too. \\ \hline

Corrections & Corrections & User provides an initial slot value and corrects or updates the value later in the conversation. \\ \hline

Adversarial Corrections & Corrections & User corrects or updates a previously specified (valid) slot value such that the the updated value is no longer valid based on constraints defined in the schema or instructions. \\ \hline

Relative Corrections & Corrections, Relative & User corrects or updates a previously specified slot in a relative manner (e.g., Actually increase the headcount by 15\% and then add 5 more). \\ \hline

Conditional Corrections & Corrections, Conditionals & User corrects a slot based on a conditional (e.g., If <condition>, then update the operating system to Linux). \\ \hline

Meta-Corrections & Corrections & User corrects a previously specified slot and then undos the correction, reverting the slot back to its original value. \\ \hline

Slot Reset & Slot Reset & User specifies a slot value and then decides to reset the slot value later in the conversation (e.g., Actually, forget what I said about <slot> ).  \\ \hline

All Slot Reset & Slot Reset & User resets all previously filled slots within the conversation. \\ \hline

Conditional Reset & Slot Reset, Conditionals & User resets a slot based on a conditional (e.g., If <condition>, then forget what I said about <slot> ). \\ \hline

Meta-Reset & Slot Reset & User resets a previously filled slot and then undos the reset, reverting the slot back to its original value. \\ \hline

\end{longtable}
\end{center}

\section{Final Joint Goal Accuracy (F-JGA)}
\label{app:fjga}
\raggedright

Let $\mathcal{D} = \{d_1, d_2, \ldots, d_N\}$ be a set of $N$ dialogues. For each dialogue $d_i$, let $S_i^* = \{(s_k, v_k)\}$ denote the ground-truth final dialogue state and $\hat{S}_i = \{(s_k, \hat{v}_k)\}$ denote the predicted final dialogue state, where $s_k$ is a slot and $v_k$ is its corresponding value. F-JGA is defined as:

\begin{equation}
    \text{F-JGA} = \frac{1}{N} \sum_{i=1}^{N} \mathbf{1}\left[\hat{S}_i = S_i^*\right]
    \label{eq:fjga}
\end{equation}

where $\mathbf{1}[\cdot]$ is the indicator function that equals $1$ if the predicted final dialogue state exactly matches the ground-truth final dialogue state, and $0$ otherwise. Unlike turn-level JGA, which computes accuracy at every dialogue turn $t$:

\begin{equation}
    \text{JGA} = \frac{1}{N} \sum_{i=1}^{N} \frac{1}{T_i}\sum_{t=1}^{T_i} \mathbf{1}\left[\hat{S}_i^t = S_i^{*t}\right]
    \label{eq:jga}
\end{equation}

F-JGA evaluates only the final turn $T_i$, making it agnostic to the number of dialogue turns and focusing solely on whether the model correctly resolves the complete dialogue state by the end of the conversation.

\section{Prompt Example}
\label{sec:prompt_example}

\begin{figure*}[t]
    \centering
    \includegraphics[height=9cm, keepaspectratio]{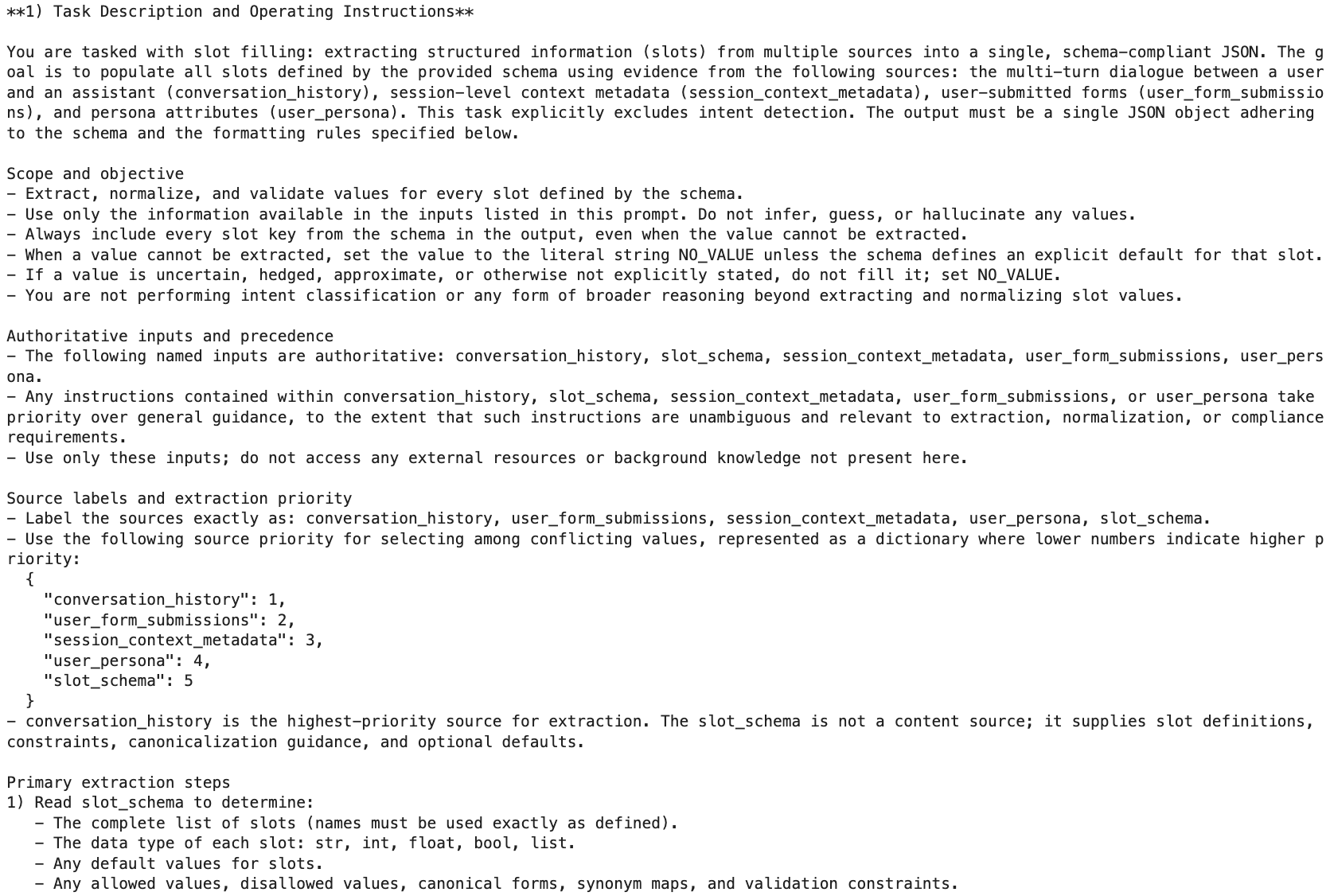}
    \includegraphics[height=9cm, keepaspectratio]{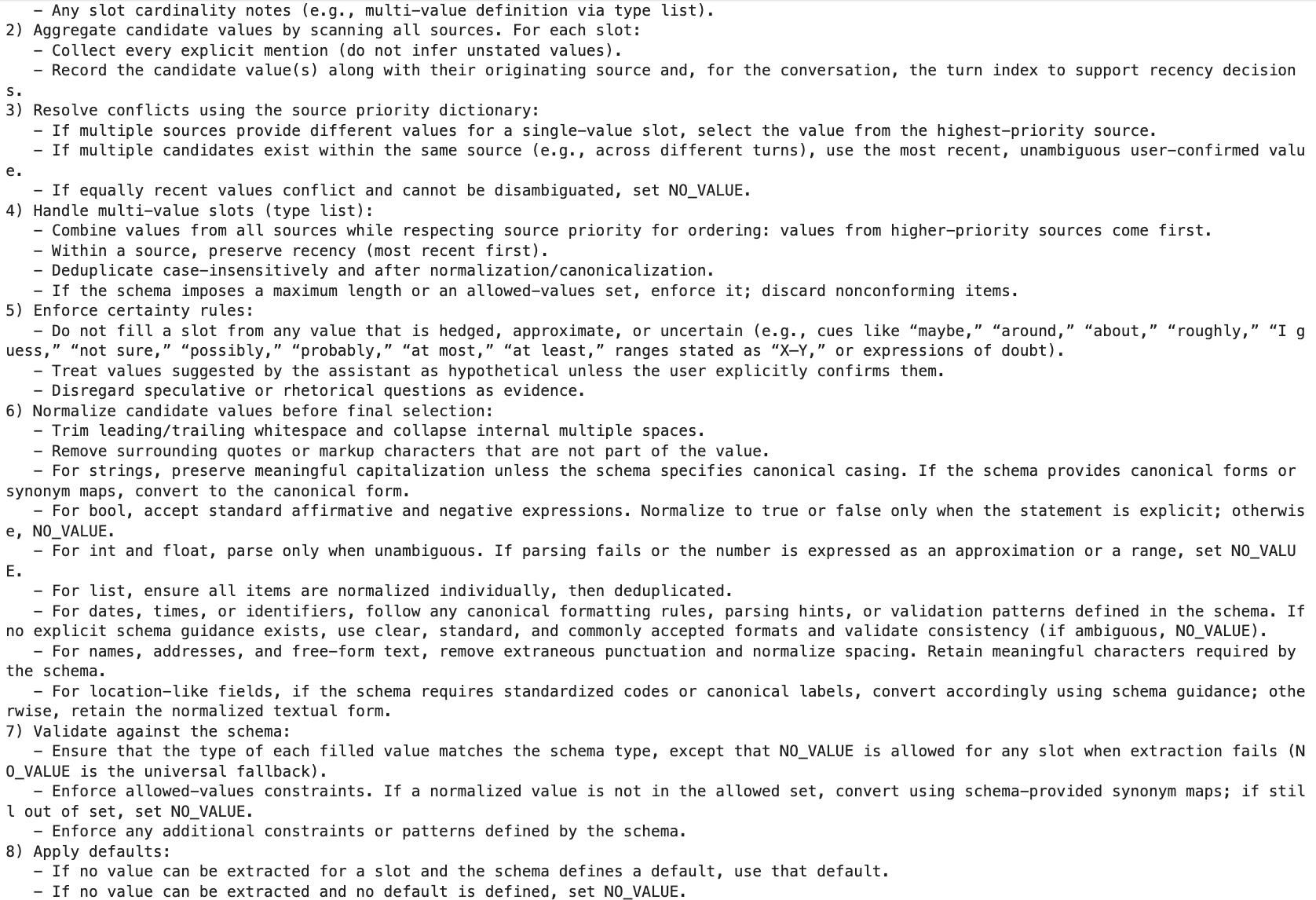}
\end{figure*}
\begin{figure*}[h]
    \centering
    \includegraphics[height=9cm, keepaspectratio]{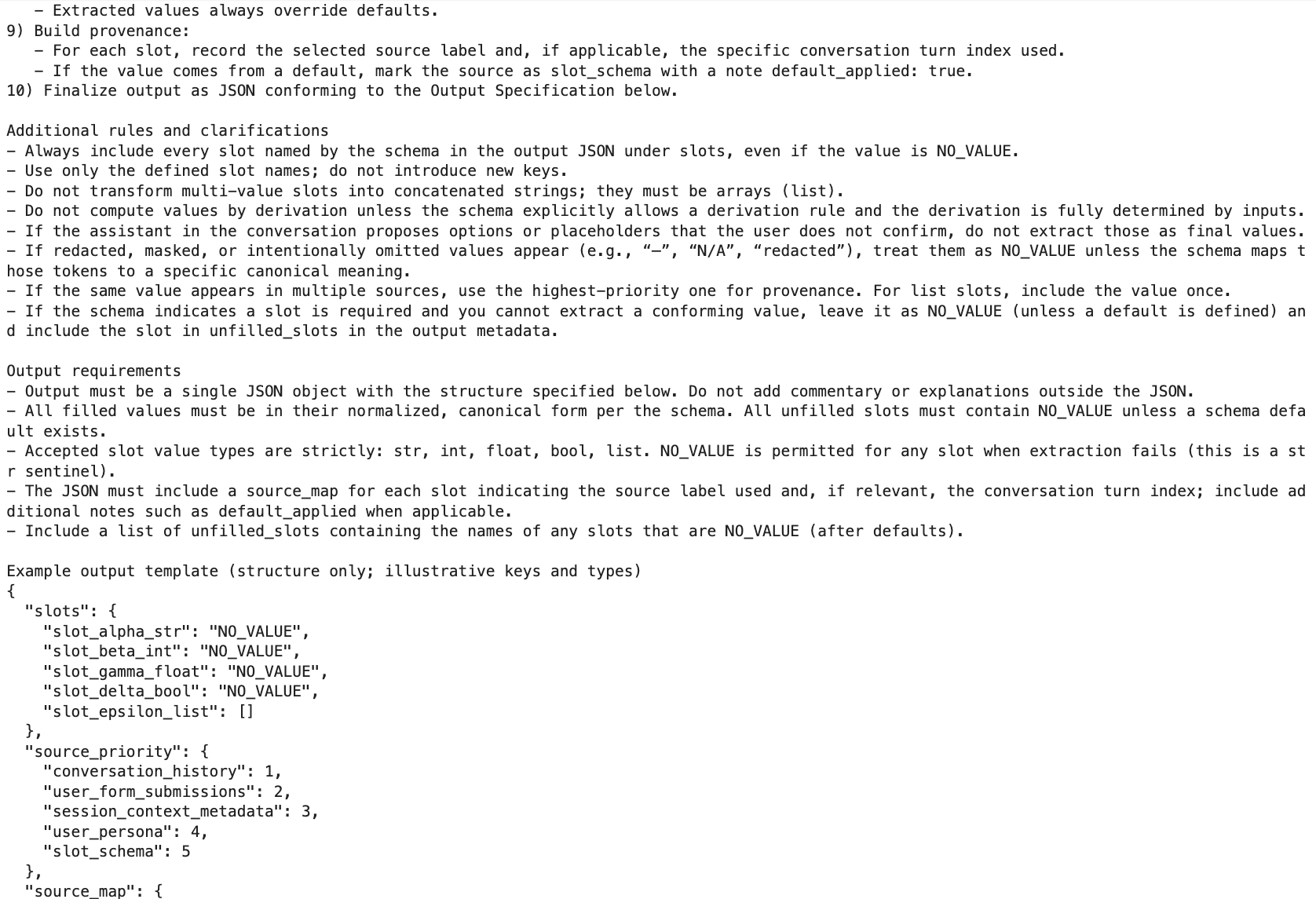}
    \includegraphics[height=9cm, keepaspectratio]{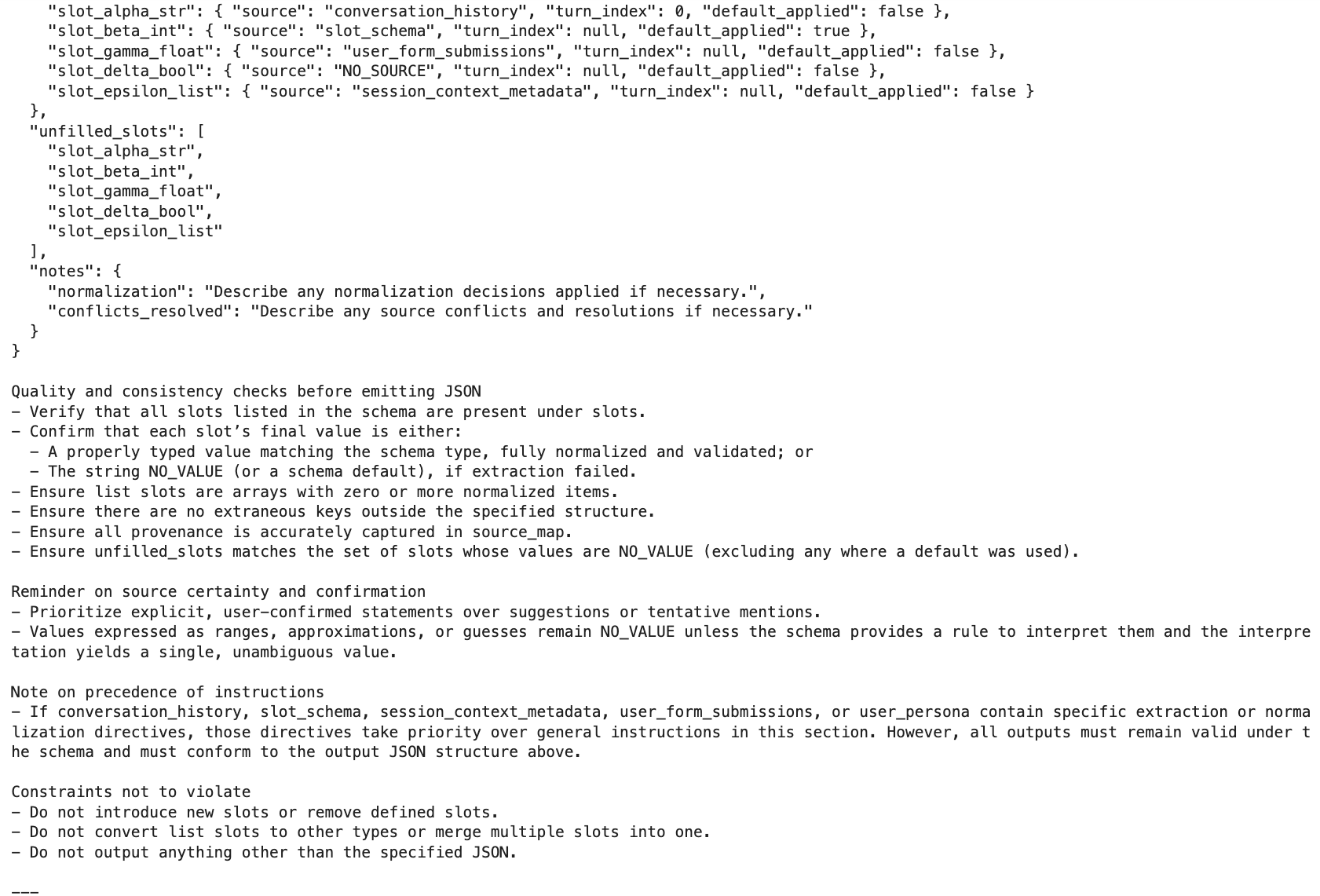}
\end{figure*}
\begin{figure*}[h]
    \centering
    \includegraphics[height=9cm, keepaspectratio]{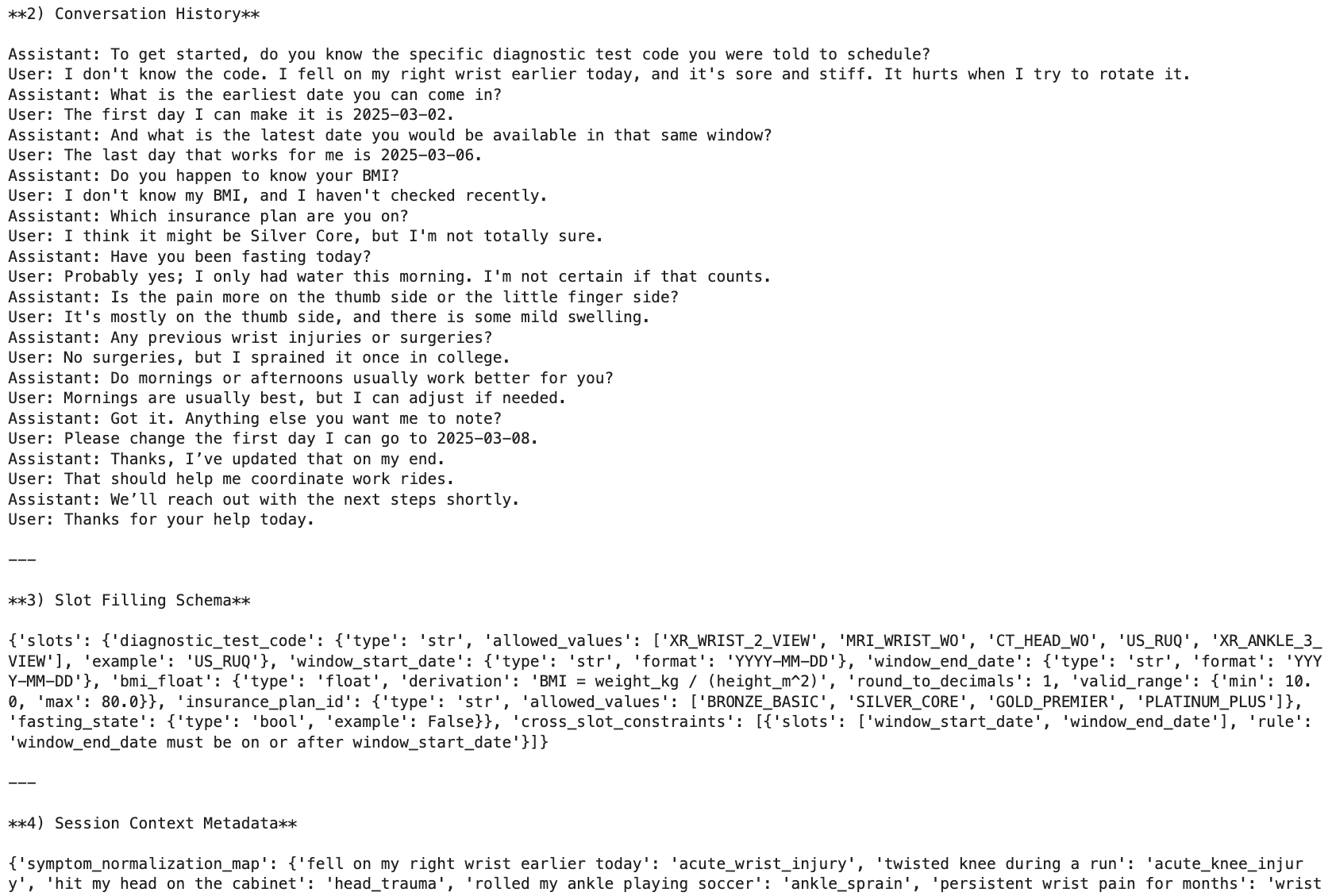}
    \includegraphics[height=4.2cm, keepaspectratio]{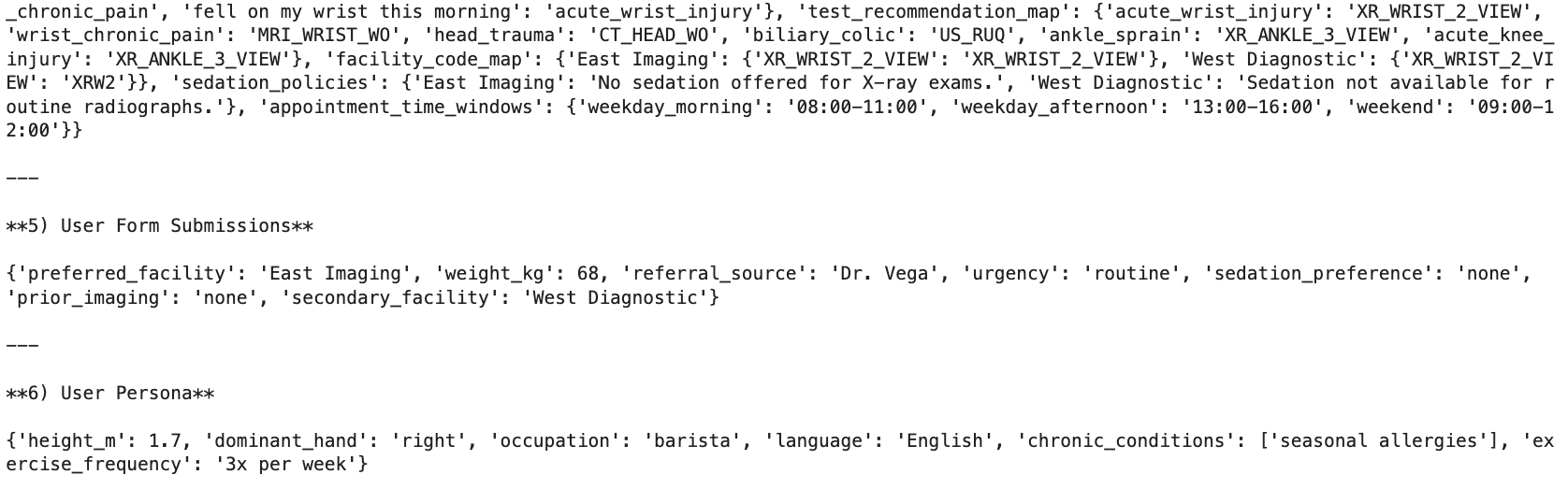}
\end{figure*}



\end{document}